\documentclass{article}

\usepackage[english]{babel}

\usepackage[letterpaper,top=2cm,bottom=2cm,left=3cm,right=3cm,marginparwidth=1.75cm]{geometry}

\usepackage{amsmath}
\usepackage{graphicx}
\usepackage[colorlinks=true, allcolors=blue]{hyperref}

\title{Depth Matching Method Based on ShapeDTW for Oil-Based Mud Imager}
\author{Li Fengfeng (email: cj21701@163.com), Feng Zhou, Wu Hongliang, \\
Zhang Hao, Tian Han, Liu Peng and Yuan Lixin\\
\\
\small PetroChina Research Institute of Petroleum Exploration and Development, Beijing 100083, China
}

\begin{document}
\maketitle

\begin{abstract}
In well logging operations using the oil-based mud (OBM) microresistivity imager, which employs an interleaved design with upper and lower pad sets, depth misalignment issues persist between the pad images even after velocity correction. This paper presents a depth matching method for borehole images based on the Shape Dynamic Time Warping (ShapeDTW) algorithm. The method extracts local shape features to construct a morphologically sensitive distance matrix, better preserving structural similarity between sequences during alignment. We implement this by employing a combined feature set of the one-dimensional Histogram of Oriented Gradients (HOG1D) and the original signal as the shape descriptor. Field test examples demonstrate that our method achieves precise alignment for images with complex textures, depth shifts, or local scaling. Furthermore, it provides a flexible framework for feature extension, allowing the integration of other descriptors tailored to specific geological features.
\end{abstract}

\section{Introduction}

\begin{figure}
    \centering
    \includegraphics[width=0.5\linewidth]{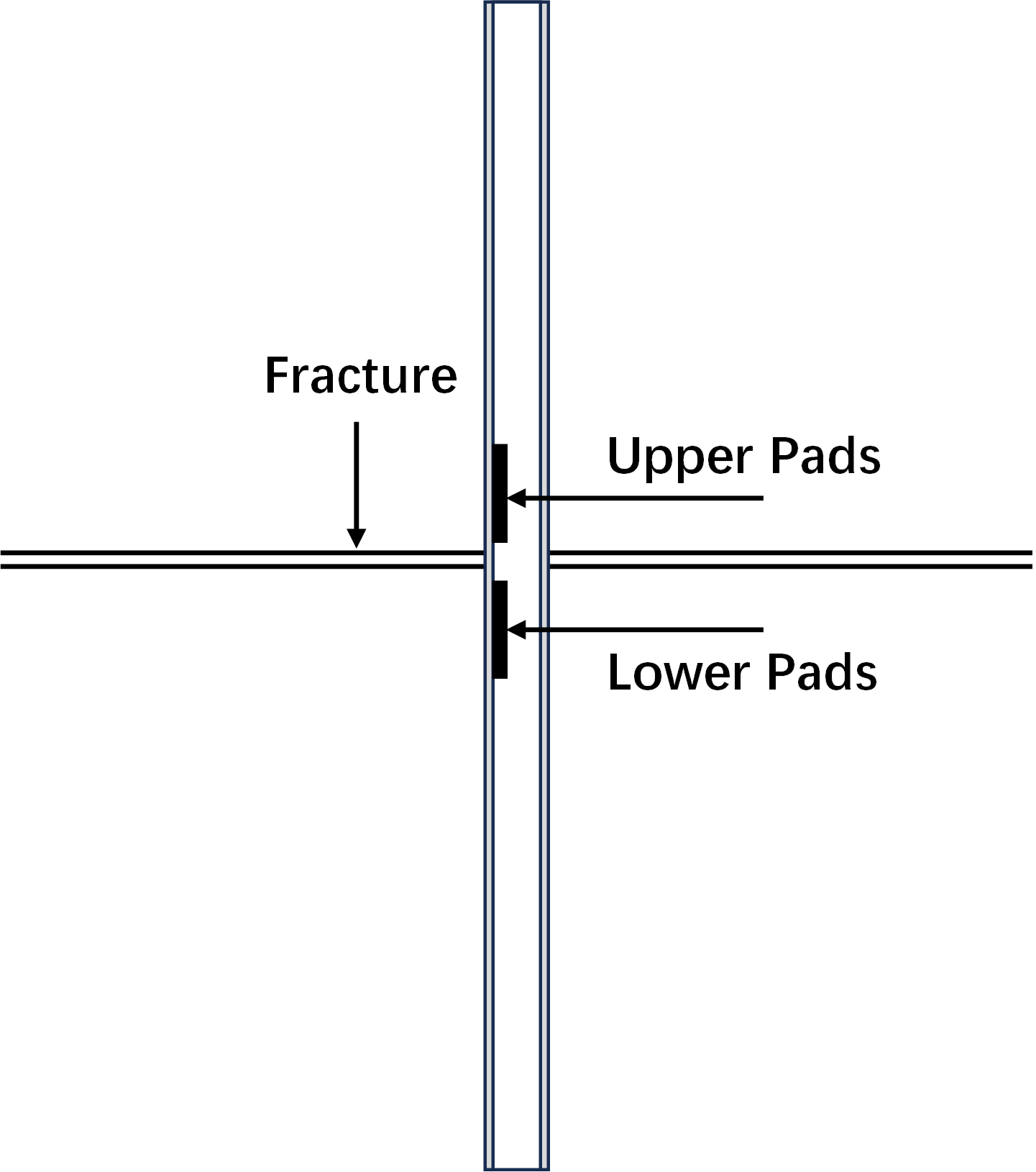}
    \caption{Schematic of fracture model and imaging pads}
    \label{fig:Pads}
\end{figure}

Fractures and vugs characterization from borehole images delivers important insights for reservoir evaluation, primarily since these structures act as the key flow networks regulating productivity in tight reservoirs \cite{dong2018fast}, \cite{ponziani2015experimental}, \cite{schlicht2020identifying}. The Formation Microresistivity Imager (FMI) provides high-resolution resistivity images of the borehole surface, delivering detailed azimuthal, sedimentary, and structural information vital for formation evaluation \cite{liu2025filling}. This data enables both the qualitative identification of features like open fractures and dissolved vugs, and the quantitative analysis of key parameters such as fracture porosity and aperture, offering reliable geophysical support for reservoir assessment \cite{jie2023characterization}. 
Oil-base mud (OBM) is preferred in challenging environments like deepwater, horizontal, and high-temperature high-pressure wells due to its superior wellbore stability and blowout prevention capabilities \cite{jinsheng2024research}. However, its nonconductive character prevents the operation of conventional DC-based FMI tools, which require a conductive water-based mud to form a current circuit \cite{chen2014inversion}. The new-generation OBM imagers operate on a capacitive coupling principle. They employ a high-frequency (MHz) alternating current to induce a displacement current through the insulating mud, thereby establishing an effective measurement loop for characterizing the formation \cite{itskovich2014improved}.
The new-generation OBM imagers use an interleaved arrangement of upper and lower pad sets (Fig.\ref{fig:Pads}) to maximize borehole coverage and enhance image quality in complex wells, including highly deviated and horizontal sections \cite{bloemenkamp2014design}. Nevertheless, despite standard velocity corrections, depth mismatches frequently persist between the pads due to factors like cable stretch and irregular pad motion. Existing solutions often involve automatically picking feature points in a reference image and applying optimization algorithms to find local depth shifts. While helpful, these methods often struggle with multi-solution problems and typically require significant manual intervention.
To overcome these limitations, this paper introduces a novel depth matching algorithm for imaging logs based on Shape Dynamic Time Warping (ShapeDTW). Our approach better preserves morphological similarity during alignment and provides a flexible framework for feature integration, presenting a robust and automated solution to the depth matching challenge.

\section{Methods}

\subsection{Dynamic Time Warping}

The Dynamic Time Warping (DTW) algorithm computes a distance measure between two temporal sequences by finding the alignment that minimizes the total alignment cost, typically based on Euclidean distance \cite{song2021application}. It allows for local non-linear warping in the time (or depth) dimension, supporting "one-to-many" or "many-to-one" mappings. This flexibility enables the alignment of sequences with unequal lengths or imperfect correspondences by matching their characteristic features as closely as possible. DTW and its variants have been successfully applied in various geophysical tasks, including depth matching of GR logs \cite{ezenkwu2023automated}, fault interpretation, seismic facies analysis\cite{song2022dynamic}, and seismic waveform alignment \cite{wang2023crosscorrelation}.

Consider two time series X = (x1, x2, …, xm) and Y = (y1, y2, …, yn). The algorithm begins by computing an m×n distance matrix D, where $D(i,j)=\parallel x_i-y_i\ \parallel$. An accumulated cost matrix C is then derived, with each element defined recursively as:
\begin{equation}
    C(i,j)=D(i,j)+min\left[C(i-1,j),C(i,j-1),C(i-1,j-1)\right]\label{1}
\end{equation}
This matrix C encodes the minimal cumulative distances. The final optimal warping path is obtained by tracing back from C(m, n) to C(1,1), following the indices that minimize the cost at each step.

\subsection{ShapeDTW}

By incorporating local shape features, the ShapeDTW algorithm advances beyond the simple point-value comparisons of traditional DTW, leading to improved preservation of morphological similarity in time series alignment \cite{zhao2018shapedtw}. The procedure involves three sequential steps: extracting local shape features, computing a feature-based distance matrix, and performing dynamic time warping.

1)	Neighborhood Subsequence Extraction: For each point in the time series, a local subsequence of radius r is extracted, centered on that point:
\begin{equation}
    subseq\left(x_i,r\right)=\left[x_{i-r},\ldots,x_i,\ldots,x_{i+r}\right]\label{2}
\end{equation}

2)Shape Feature Extraction: A shape descriptor (e.g., raw subsequence, slope, or extrema) is applied to each extracted subsequence to encode its local structural characteristics:
\begin{equation}
    Feature\left(x_i\right)=\left[\phi_1(subseq\left(x_i,r\right)),\phi_2(subseq\left(x_i,r\right)),\ldots\right]\label{3}
\end{equation}
Where, $\phi(\cdot)$ is the feature extraction function.

3) Calculation of feature distance: The point-wise distance matrix is built by calculating the Euclidean distance between the feature vectors of the subsequences of each series:
\begin{equation}
    D_{\mathrm{shape}}(i,j)=\parallel Feature(x_i)-Feature(y_i)\ \parallel\label{4}
\end{equation}

4)Path Optimization: Standard DTW is performed on the resulting feature distance matrix to find the optimal alignment path.

\subsection{HOG-1D}

The Histogram of Oriented Gradients (HOG) descriptor characterizes object shape and edge information by capturing the distribution of local gradient orientations within an image, and was initially developed for pedestrian detection \cite{dalal2005histograms}. HOG-1D extends this concept to one-dimensional time series signals, extracting local structural features through a four-step process: gradient computation, orientation binning, cell histogram construction, and block normalization \cite{zhao2018shapedtw}.

The HOG-1D descriptor is defined by three key parameters: CellSize, BlockSize, and NumBins. The CellSize determines the number of sample points within each local cell, thereby controlling the granularity and spatial resolution of the feature extraction. The BlockSize specifies the number of consecutive cells grouped into a block, which governs the contextual scope for normalization. The NumBins parameter sets the number of orientation bins, defining the angular resolution.

\begin{figure}[!htbp]
    \centering
    \includegraphics[width=1\linewidth]{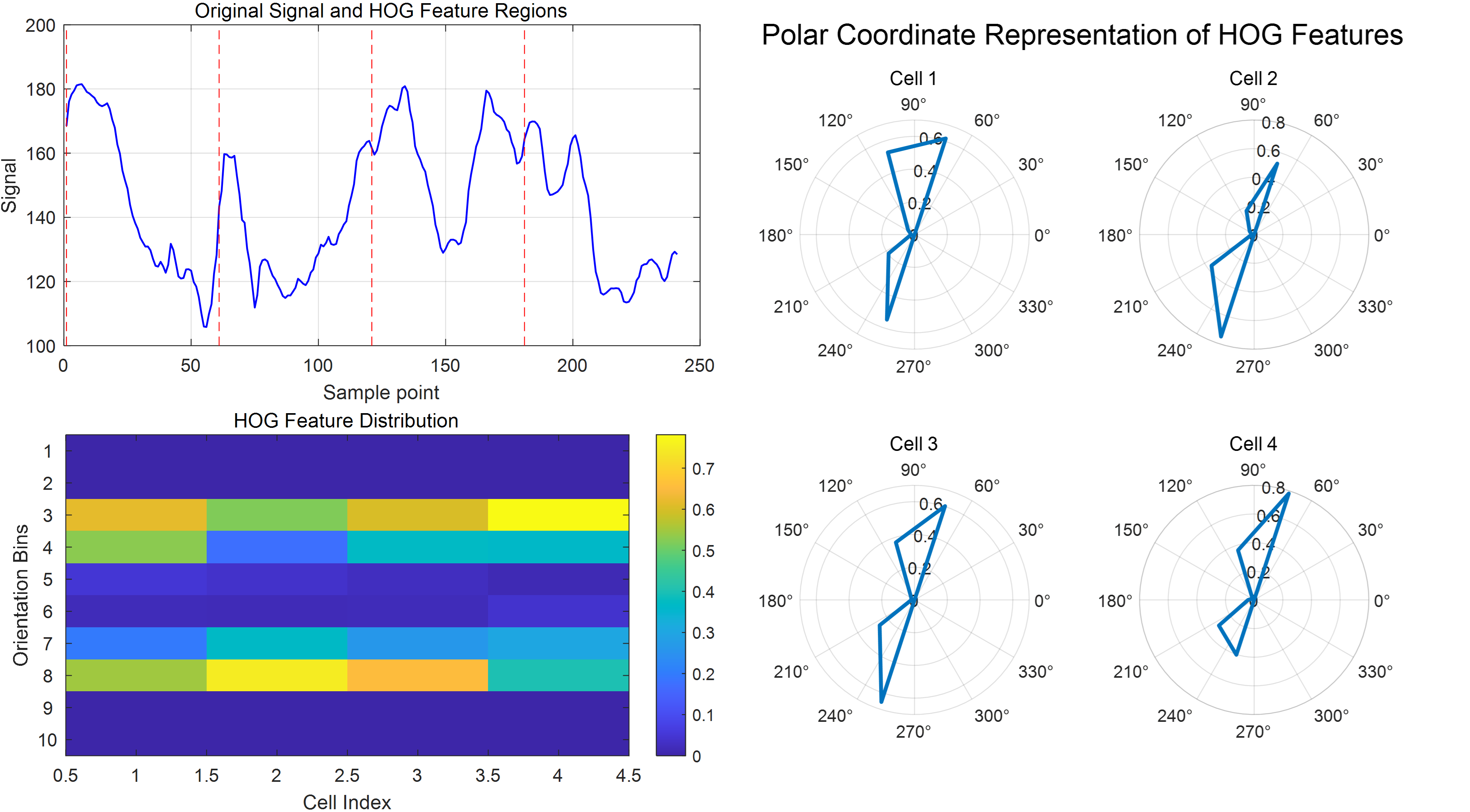}
    \caption{Results of HOG-1D feature extraction}
    \label{fig:2}
\end{figure}

In this study, the parameters are set as follows: CellSize = 60, NumBins = 10, and BlockSize = 2. Fig.\ref{fig:2} illustrates how HOG-1D captures local gradient information from a signal. The original signal is divided into 4 cells (x-axis of the heatmap), and the orientation space is quantized into 10 bins (y-axis of the heatmap). With every 2 consecutive cells forming one block, a total of 6 blocks are created. The heatmap visualizes the weighted distribution across orientation bins for each cell, while the polar plots associated with each cell depict the distribution of gradient directions within that cell.

HOG-1D is employed as the shape descriptor to extract local features from electrical imaging signals and enhance the depth-matching capability of the DTW algorithm.

\subsection{The workflow of depth matching}
\begin{figure}[!htbp]
    \centering
    \includegraphics[width=0.8\linewidth]{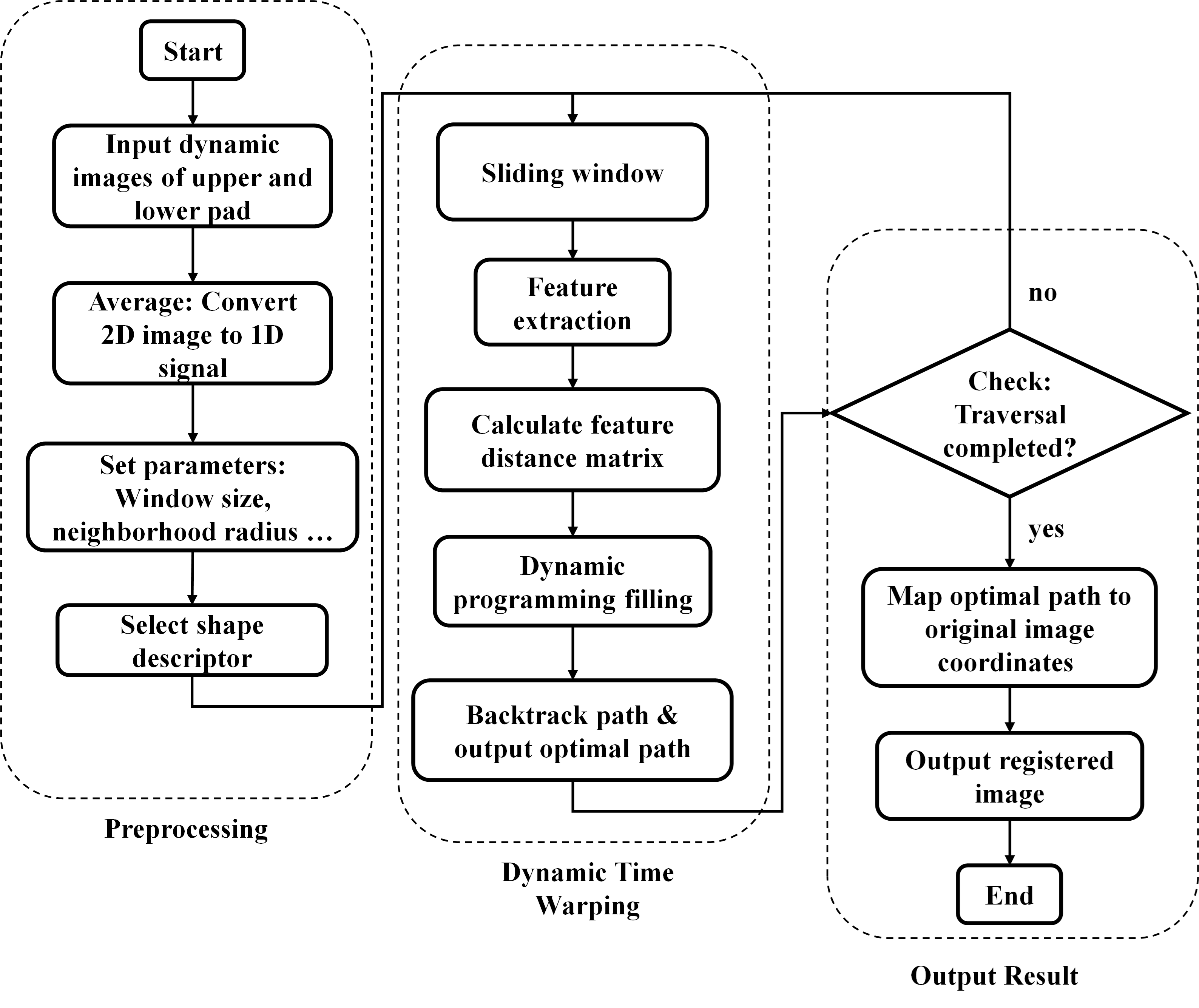}
    \caption{Depth matching workflow}
    \label{fig:3}
\end{figure}

The depth matching procedure, outlined in Fig.\ref{fig:3}, utilizes the dynamic images from the upper (reference) and lower (target) pads of the OBM borehole imager. 

First, the 2D log image undergoes pre-processing. The measurements from the upper and lower pads at each depth are averaged, converting the 2D registration challenge into a more tractable 1D signal alignment problem.

Subsequently, to manage the high computational and memory demands of DTW caused by the high-density sampling of the log, the data is processed in batches using a sliding window of 10-30 ft.

Within each window, the 1D signals from both pads are input into the ShapeDTW algorithm—equipped with a suitable shape descriptor—to determine the local optimal warping path. Finally, after iterating through all windows, the composite displacement path for the lower pad is applied to align it with the upper pad's depth domain, achieving the final depth match.


\section{Field data examples}
\begin{figure}[!htbp]
    \centering
    \includegraphics[width=1\linewidth]{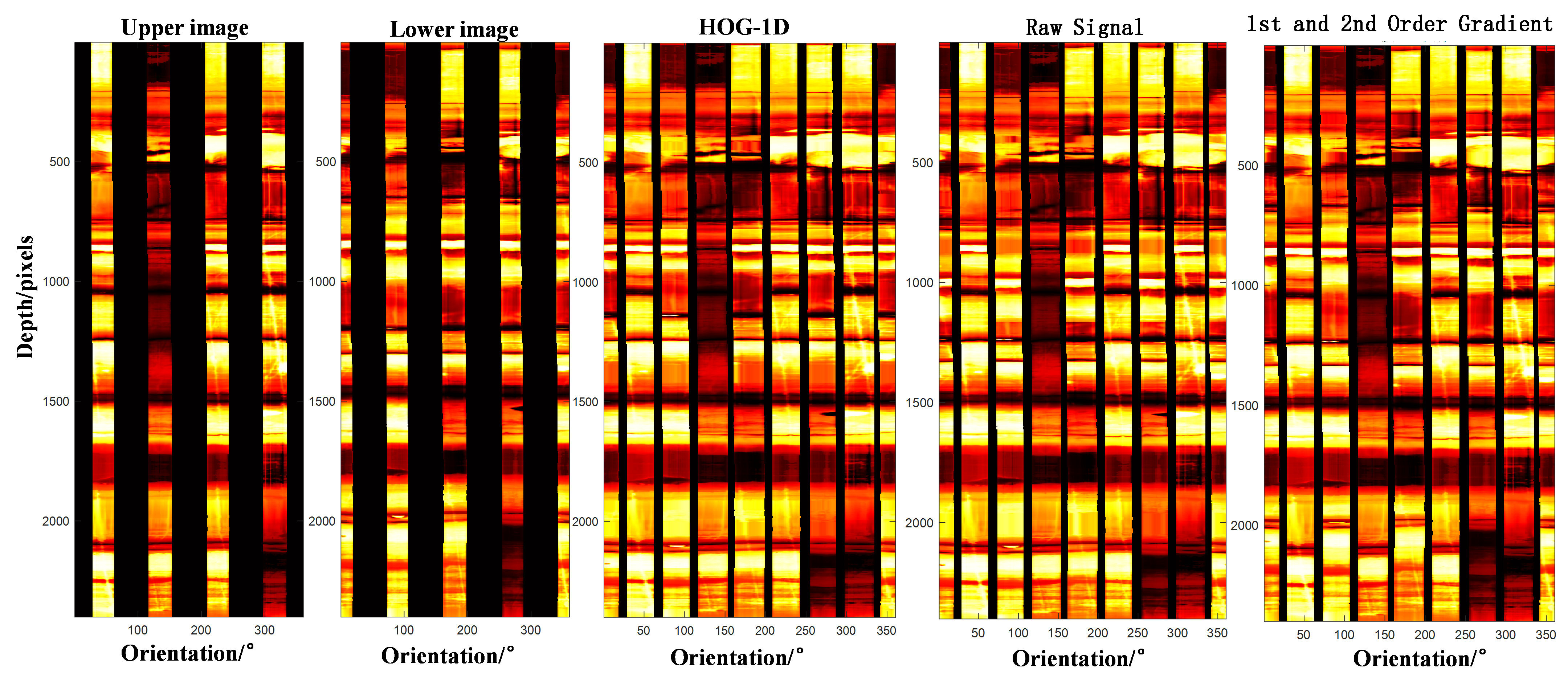}
    \caption{Depth matching results using different shape descriptors}
    \label{fig:4}
\end{figure}

Fig.\ref{fig:4} shows the depth matching results using different shape descriptors for a 20-ft log section. The figure includes: the original dynamic images from the upper and lower pads without depth matching; the matching result using HOG-1D alone as the shape descriptor; the result using the raw signal alone; and the result using a combined descriptor of first and second-order gradient signals. A comparative analysis reveals that HOG-1D as the shape descriptor exhibits the best performance, with only one fracture failing to align correctly. Using the raw signal leads to mismatches in three textural features, while the combination of first and second-order gradients yields the poorest matching outcome. Consequently, the combined features of the HOG-1D descriptor and the raw signal are selected as the shape descriptor for subsequent analysis.

\begin{figure}[!htbp]
    \centering
    \includegraphics[width=1\linewidth]{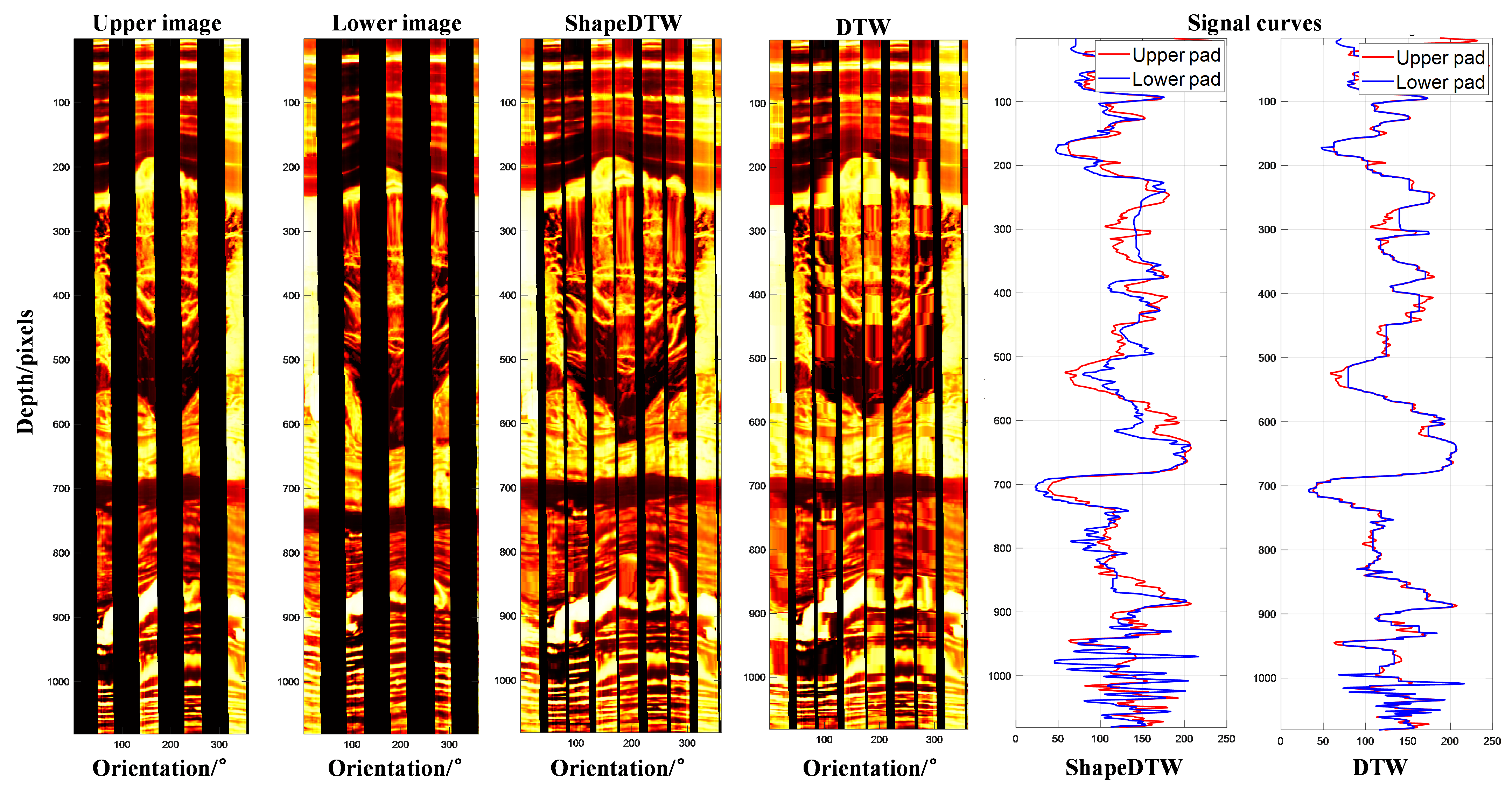}
    \caption{Comparison of depth matching results using DTW and ShapeDTW}
    \label{fig:5}
\end{figure}

Fig.\ref{fig:5} compares the depth matching results of the DTW and ShapeDTW algorithms on a 9-ft image section. The figure displays the original pad images, the registered images, and a comparison of the corresponding 1D signals. The results indicate that the DTW algorithm suffers from pronounced overfitting, evidenced by an excessive stretching of the 1D signal that distorts the original shape of the image. In contrast, the ShapeDTW algorithm successfully preserves the morphological characteristics of the original image and delivers a stable depth matching result.

\begin{figure}[!htbp]
    \centering
    \includegraphics[width=1\linewidth]{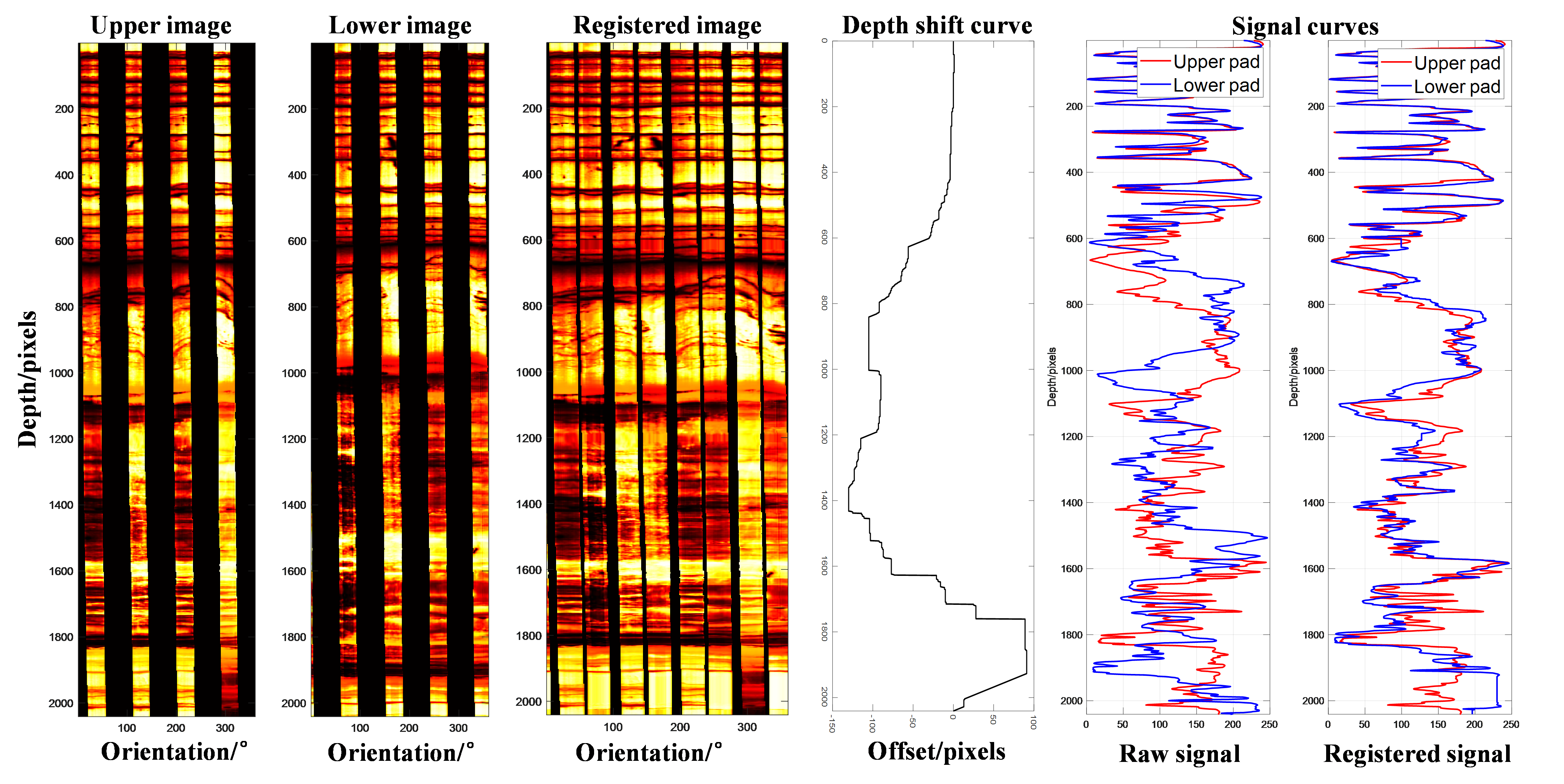}
    \caption{Results of depth matching for a 17-ft borehole image section}
    \label{fig:6}
\end{figure}

Fig.\ref{fig:6} demonstrate the application of the ShapeDTW algorithm for depth matching on log sections of 17 ft in length. In addition to the image comparisons, this figure include a depth shift curve. The results show that the proposed algorithm competently accomplishes the depth matching task even for complex borehole imaging logs with intricate textures, significant depth shifts, or substantial local image scaling.


\section{Conclusion}
The conventional DTW algorithm performs nonlinear sequence alignment by minimizing cumulative distance. However, its dependence on point-wise Euclidean distance causes it to ignore crucial shape information, resulting in image distortion when used for depth matching in oil-based mud imager data. We adopt the ShapeDTW algorithm, which utilizes shape descriptors to encode the morphological features of local subsequences. This method ensures the preservation of global morphological similarity and allows for adaptable feature expansion.

The performance of various shape descriptors—HOG-1D, raw sequence, and first/second-order gradients—was systematically evaluated. Results identify HOG-1D as the most effective in capturing discriminative features. The combination of HOG-1D and raw sequence features was empirically validated across multiple well intervals. It consistently delivers competent depth matching performance, even for challenging resistivity images characterized by complex textures and substantial local scaling or deformation.

\bibliographystyle{ieeetr}
\bibliography{ref}

\end{document}